\def\year{2022}\relax
\documentclass[letterpaper]{article} % DO NOT CHANGE THIS
\usepackage{aaai22}  % DO NOT CHANGE THIS
\usepackage{times}  % DO NOT CHANGE THIS
\usepackage{helvet}  % DO NOT CHANGE THIS
\usepackage{courier}  % DO NOT CHANGE THIS
\usepackage[hyphens]{url}  % DO NOT CHANGE THIS
\usepackage{graphicx} % DO NOT CHANGE THIS
\usepackage{natbib}  % DO NOT CHANGE THIS AND DO NOT ADD ANY OPTIONS TO IT
\usepackage{caption} % DO NOT CHANGE THIS AND DO NOT ADD ANY OPTIONS TO IT
\DeclareCaptionStyle{ruled}{labelfont=normalfont,labelsep=colon,strut=off} % DO NOT CHANGE THIS
\usepackage{algorithmic}

\usepackage{newfloat}
\usepackage{listings}
\usepackage{soul}
\usepackage{amsmath}
\usepackage{url}
\usepackage[ruled]{algorithm2e}

\SetKwInOut{Parameter}{parameter}
\SetKwData{Left}{left}
\SetKwData{This}{this}
\SetKwData{Up}{up}
\SetKwFunction{Union}{Union}
\SetKwFunction{FindCompress}{FindCompress}
\SetKwInOut{Input}{input}
\SetKwInOut{Output}{output}
\SetKwInput{KwInput}{Input}                % Set the Input
\SetKwInput{KwOutput}{Output} 

\usepackage{booktabs,subcaption,amsfonts,dcolumn}

\usepackage{multirow}

\author {
    Vinod K. Kurmi,\thanks{Equal contributions.}\textsuperscript{\rm 1}
    Rishabh Sharma,\footnotemark[1]\textsuperscript{\rm 2}
    Yash Vardhan Sharma,\footnotemark[1]\textsuperscript{\rm 2}
   Vinay P Namboodiri\textsuperscript{\rm 3}
}
\affiliations {
    \textsuperscript{\rm 1} KU Leuven, Belgium\thanks{Work done at IIT Kanpur, India},  \textsuperscript{\rm 2} IIT Roorkee, India
    \textsuperscript{\rm 3} University of Bath, UK\\
    vinod.kurmi@kuleuven.be,  \{rsharma1, ysharma\}@me.iitr.ac.in, vpn22@bath.ac.uk
}

\title{Gradient Based Activations for Accurate Bias-Free Learning}

\begin{document}

\maketitle

\begin{abstract}
Bias mitigation in machine learning models is imperative, yet challenging. While several approaches have been proposed, one view towards mitigating bias is through adversarial learning. A discriminator is used to identify the bias attributes such as gender, age or race in question.  This discriminator is used adversarially to ensure that it cannot distinguish the bias attributes.  The main drawback in such a model is that it directly introduces a trade-off with accuracy as the features that the discriminator deems to be sensitive for discrimination of bias could be correlated with classification. In this work we solve the problem. We show that a biased discriminator can actually be used to improve this bias-accuracy tradeoff. Specifically, this is achieved by using a feature masking approach using the discriminator's gradients. We ensure that the features favoured for the bias discrimination are de-emphasized and the unbiased features are enhanced during classification. We show that this simple approach works well to reduce bias as well as improve accuracy significantly. We evaluate the proposed model on standard benchmarks. We improve the accuracy of the adversarial methods while maintaining or even improving the unbiasness and also outperform several other recent methods.

% \textbf{Through our approach we promote learning through unbiased and non-stereotypical feature representation along with ensuring that features favoured for bias discrimination are de-emphasized during classification.}

% features favoured for classification are damped while discriminating bias and 

\end{abstract}
\section{Introduction}
\label{intro}
The issue of bias in computer vision has been widely studied where the bias could be in terms of under-represented class samples \cite{li2019repair}, gender \cite{wang2019balanced}, demographics \cite{lahoti2020fairness} or other cases. %to add more references. Check Olga's CVPR 2020 paper
The use of computer vision has a variety of practical applications ranging from autonomous driving to medicine. Particularly the use of vision systems in applications such as face recognition \cite{robinson2020face} and image generation \cite{ramaswamy2020fair} are widespread. The bias in such systems may unduly affect these systems. Practical instances have been observed as that of image super-resolution ~\cite{menon2020pulse} resulting in generating white race images for down-sampled images from other races. %Similarly an image cropping algorithm used by Twitter showed that it favoured white race people while cropping \footnote{https://blog.twitter.com/en\_us/topics/product/2020/transparency-image-cropping.html}. 

\begin{figure*}[h!]
\centering
\begin{subfigure}{0.33\textwidth}
  \centering
  \includegraphics[width=1.0\linewidth]{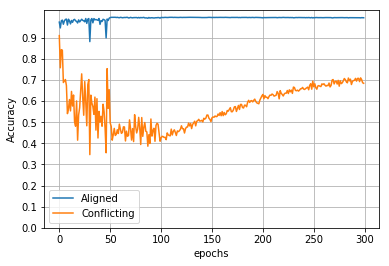}
  \caption{Discriminator on the features of a vanilla model}
  \label{fig:disc_erm}
\end{subfigure}
\begin{subfigure}{0.33\textwidth}
  \centering
  \includegraphics[width=1.0\linewidth]{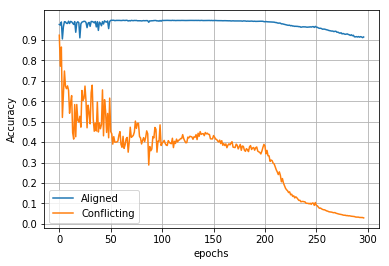}
  \caption{Discriminator on the features of an adversarially trained model}
  \label{fig:disc_adv}
\end{subfigure}
\begin{subfigure}{0.33\textwidth}
  \centering
  \includegraphics[width=1.0\linewidth]{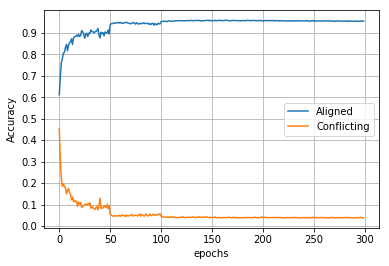}
  \caption{Discriminator on the features of an adversarially trained model with GBA (ours)}
  \label{fig:disc_gba}
\end{subfigure}

\caption{Accuracy plots of discriminator on aligned and conflicting samples on different methods}
\vspace{-1em}
\end{figure*}

In this paper, we are particularly concerned by such bias in computer vision and would like to address this.

One class of approaches used to solve this problem is to use an adversarial learning technique \cite{louppe2017learning,madras2018learning} where we use a discriminator to distinguish the particular attributes such as race, gender or age that we are concerned about. This discriminator is trained in an adversarial learning framework  in a manner reminiscent of the domain adaptation approach \cite{ganin2015unsupervised}. That is, the loss from the discriminator is reversed while training the `feature extractor'. This is achieved by taking a set of deep learning layers to form a feature extractor and branching from this to two networks a classifier and a discriminator. The classifier thus receives features through which at the end of training, a discriminator is not able to distinguish a bias attribute in question (race/gender). The idea is appealing and can be effective in mitigating bias. However, a drawback that is introduced through this framework is that it directly introduces a trade-off with accuracy. This is particularly the case when there is high skew in terms of the particular attributes, i.e. gender or race, then the accuracy suffers significantly. In our work, we use a simple method to ensure that we obtain both objectives, that is, the bias is minimized and accuracy is high.\\
The crux of our idea is to ensure explicit de-correlation between the adversarial discriminator and classifier. This is because, when there is substantial skew in the data, for instance very few images of a specific attribute (gender or race for instance) then the discriminator has very few samples to distinguish the set of `bias' attributes (the set of attributes we are concerned should not be discriminated against). As a result, the discriminator learns more through the set of features that are related to the classifier's set of labels i.e. in the output space of the classifier. For instance, if we consider we are distinguishing digits between 0 to 9 and almost all the samples are grey samples and very few samples are colored samples, (for instance samples of 4 and 6 only) then the discriminator tries to identify the class label in an effort to distinguish whether the sample is colored or not. Thus, in the example knowing whether the sample is 4 or 6 would help the discriminator in predicting whether the digits are colored or not. This is particularly the case as we are using the discriminators loss adversarially and ensuring that the discriminator has a hard time ensuring whether the digits are colored or not.
 In this paper, we propose to use a biased discriminator to improve the accuracy of adversarial methods while debiasing the feature representation. We use gradients of the discriminator and propose a masking scheme for the features, we term as Gradient Based Activation (GBA). Use of GBA provides us a masking rule to drop certain features in order to do unbiased training. The effectiveness of this approach is demonstrated and validated.
We further describe the proposed approach in detail in coming sections and provide its extensive analysis to demonstrate the efficacy of the proposed method.

\section{Related Work}
\textbf{Bias in Computer Vision}\\
The issue of bias in computer vision has been considered by several works~\cite{grover2019bias,kim2019learning,quadrianto2019discovering,ganin2015unsupervised}. It causes unfair or biased leaning in computer vision tasks such as face recognition~\cite{robinson2020face,xu2020investigating}, object detection~\cite{de2019does} and image generation~\cite{xu2018fairgan}. Data imbalance is one of the source of bias learning~\cite{buolamwini2018gender}. Ramaswamy \textit{et al}~\cite{ramaswamy2020fair} tackle it by generating realistic samples from the GANs. It has been shown that balanced data also faces bias in feature representations~\cite{wang2019balanced}.

\noindent\textbf{Adversarial debiasing approaches}\\
Some of the works that pursued adversarial learning include the works by \cite{louppe2017learning,kurmi2019attending} and \cite{madras2018learning}. Similarly, to prevent gender bias, Wadsworth \textit{et al.}~\cite{wadsworth2018achieving} present an adversarially-trained neural network that predicts recidivism and is trained to remove racial bias using a discriminator. \cite{adel2019one,zhang2018mitigating} includes a new hidden layer to enable the concurrent adversarial optimization for fairness and accuracy. Adversarial bias removal methods are also applied in text data.~\cite{elazar2018adversarial}. In recent work~\cite{lahoti2020fairness} train an adversarial reweighting approach for improving fairness.

\noindent\textbf{Other debiasing efforts}\\
 A number of other approaches have also been considered for solving this problem. In \cite{wang2019balanced}, the authors show that even when datasets are balanced (each label co-occurs equally with each gender), learned models do amplify the association between labels and gender. \cite{kiritchenko2018examining,vig2020investigating} analyze the gender bias in the case of sentiment analysis. Another gender bias work~\cite{buolamwini2018gender} evaluates bias present in automated facial analysis algorithms and datasets for phenotypic subgroups.  
\cite{leino2018feature} demonstrates that bias amplification can arise via an inductive bias in gradient descent methods.   A kernel density estimation~\cite{cho2020fair} is applied to tackle the fairness problem. Gat {\it et al.} \cite{gat2020removing} present a regularization term based on the functional entropy to remove a classifier's bias. In another work, \cite{nam2020learning} show that sample performance-based methods can be used to avoid the bias in the model.  A  disentanglement approach to obtain the bias invariant representation has been presented in~\cite{sarhan2020fairness}. \\
In contrast to these other techniques, our work is focused on solving the drawback that we identify in adversarial learning framework for debiasing. Further, our work provides insight into the role of feature representations and feature masking while training a classifier and informs us about the source of bias in a classifier.
% Generalization paper

% \section{Biased Discriminator}
% Adversarial learning based methods mitigate the bias by introducing a discriminator to learn adversarial features to the protected variables or domain. This can be obtain by a gradient reverse layer or a min-max optimization of feature extractor and discriminator. 

\section{Bias in Machine Learning}
\label{mot}
In any dataset we may have classes that are highly skewed towards a particular sensitive attribute. This skew leads to a classifier correlating class information with these sensitive attributes and hence inducing stereotypes in learning. When the sensitive attributes are easy to learn, there will be no incentive for a model to learn class features. For instance, in a dataset, if images of horse are dominantly in color (RGB channels exist) and that of deer are dominantly grey (only grey channel exists), a grey horse, may be misclassified as a deer based on the number of channels it has, rather than the appearance. This would be because, the classifier would associate `greyscale' attribute with the label `deer'.
For this example, all examples of `colored horses' and `greyscale deer' are aligned with the bias that exists in the dataset and we term these as `bias aligned' samples. The examples of `greyscale horses' and `colored deer' are termed as `bias conflicting' samples as they are not following the dominant bias in the dataset.\\
In the setting of bias/fairness we want a classifier to be agnostic to these bias attributes. A discriminator model trained on the features of the classifier and its ability to discriminate among the bias attributes aims to give us a measure of bias in the feature representation. For instance, the discriminator for the example would aim to classify whether a feature representation is of color images or that of greyscale images. This discriminator used adversarially would aim to make the feature representation invariant to the greyscale/color attribute. In general feature space is entangled with bias and class information. The purpose of the debiasing approaches is to remove the entangled bias information from the feature space.\\
Adversarial approaches build on the above hypothesis and use an adversarial loss in order to debias the feature space. But it has been seen and stated in several works~\cite{wadsworth2018achieving,madras2018learning} that adversarial approaches for debiasing introduce a trade-off with accuracy.\\
In this work we identify the reason for this trade-off with empirical justifications and provide a simple and effective method to address this issue.

\subsection{Class correlation of discriminator}

In Fig \ref{fig:disc_erm} we provide a discriminator's validation accuracy plot for a vanilla model on the CIFAR-10S dataset. We see a discrepancy in the performance between bias-aligned and conflicting samples. In the case of CIFAR-10S, the samples are biased based on the color attribute. This clearly indicates that a discriminator benefits from class correlations in its prediction i.e. a sample of a given bias in its dominant class is identified with higher accuracy by the discriminator due to the presence of class correlation with the bias. This makes the discriminator itself biased. It has high accuracy for the bias aligned samples. For instance for some classes, on the basis of alignment with the dominant color attribute, the discriminator performs well. On the other hand this type of class correlation for predicting the bias attribute will harm the discriminator accuracy on bias conflicting samples. For these samples a discriminator will perform well if and only if there is certain bias information in the representation. That is, for the example `greyscale horse', the discriminator performs badly. It will perform well for these samples only if the discriminator actually represents the color information accurately. This implies discriminator performance on bias conflicting samples is the indicator of bias rather than on the whole test set. 
% And when a discriminator is given a representation without having any bias information, it will correlate the class with the bias attribute to achieve perfect accuracy on bias aligned samples, while giving wrong predictions for bias conflicting samples.

\subsection{Adversarial learning}

% A class of methods use the adversarial loss of discriminator to debias the feature representation. In Fig \ref{fig:disc_adv} we show the discriminator's validation accuracy plot for the adversarial method.

%Here we observe adversarial training degrades the accuracy on bias aligned samples, along with the bias conflicting samples. As the bias aligned samples correlate with the true class, this implies that the class correlation of the samples is reduced. This would result also in a decrease in accuracy for the bias conflicting samples as we show later.  %and hence debiasing the representation. 

% Being a biased discriminator it correlates with class features and hence the bias aligned accuracy can be maintained high. However, here we observe that in order to debias, the adversarial loss depreciates the bias aligned accuracy along with the bias conflicting accuracy hence degrading the class correlated features in the process of debias causing a bias-accuracy trade-off. This can be observed by considering also the classifier's accuracy for the adversarial training. We show that through adversarial training, while the discriminator's accuracy reduces, the classifier's accuracy also reduces. This is due to the correlation with bias. This bias-accuracy trade-off is undesirable.

A class of methods use the adversarial loss of discriminator to debias the feature representation. Adversarial loss is implemented to cause degradation in the discriminator's performance, as we discussed in the previous section discriminator may use class cues for the discrimination and simultaneously there is a classifier which will try to learn these class cues. So in a sense we have a conflicting objective here which may harm the class features rather than the bias. We show the discriminator's validation accuracy plot in Fig \ref{fig:disc_adv}, being a biased discriminator it correlates with class features and hence the bias aligned accuracy can be maintained high. However, here we observe that in order to debias, the adversarial loss depreciates the bias aligned accuracy along with the bias conflicting accuracy hence degrading the class correlated features in the process of debias causing a bias-accuracy trade-off. This can be observed by considering also the classifier's accuracy for the adversarial training as we show later. We show that through adversarial training, while the discriminator's accuracy reduces, the classifier's accuracy also reduces. This is due to the correlation with bias. This bias-accuracy trade-off is undesirable.

% which also highlights how use of this sub-optimal objective causes bias-accuracy trade-off i.e. in order to decrease the bias conflicting accuracy it also depreciate the bias aligned accuracy. This motivates us to stop the signals from bias correlated class features from getting into the objective.

% This is because as the bias information present in the feature representation decreased, the discriminator's bias increases. 

A question may arise that why is bias conflicting accuracy even worse than random? This is because more debias the classifier more biased is the discriminator 
% (or This is because as the bias information present in the feature representation decreased, the discriminator's bias increases. )
, i.e. it correlates more with the class cues, hence it predicts the bias attribute by associating with the class of the sample and as the bias conflicting samples don't follow the dominant bias hence it predicts the wrong attribute. This also means that the correlated features in this case will be bias free. 
Using these observations we motivate the proposed approach in the next section.

\subsection{Motivation}

In the above sections we have discussed about the behaviour of discriminator on different samples, the possibility of it being biased and how adversarial loss from such a discriminator is responsible for the bias-accuracy trade-off.\\
We started with the problem of bias in the classifier and ended up having a biased discriminator. We propose how a biased discriminator can rather be used as an effective tool for debiasing, which can by itself prevent the bias-accuracy trade-off as well as promote an unbiased feature representation.

We analyse this carefully and obtain a method in this paper that shows how a biased discriminator can be used to debias the classifier without compromising on accuracy in an adversarial framework. 

The use of biased discriminator is based on the following observation: 
\begin{itemize}
    \item When the prediction of a discriminator is correct, it is attending to the features correlated with the bias attribute. Such features are unwanted in our representation, hence masking them during the classifier training will encourage the classifier to learn through the unbiased features. This masking will also prevent the adversarial loss from degrading the class features correlated with the bias.
    \item 
    In the case when the discriminator's prediction is incorrect, it is attending to the features with no bias attribute information and rather is spuriously correlating features with the predicted incorrect bias attribute. For instance, it is predicting `greyscale' just by observing a deer. In this case, the features are correlated with the class. Hence, we propose to enhance these features to promote unbiased learning in our classifier. In this case, as the discriminator is not able to predict the right bias attribute, it implies that the learning will be neutral with respect to the bias attributes.
\end{itemize}

\begin{figure*}
 \centering
    \includegraphics[height=5cm,width=13.5cm]{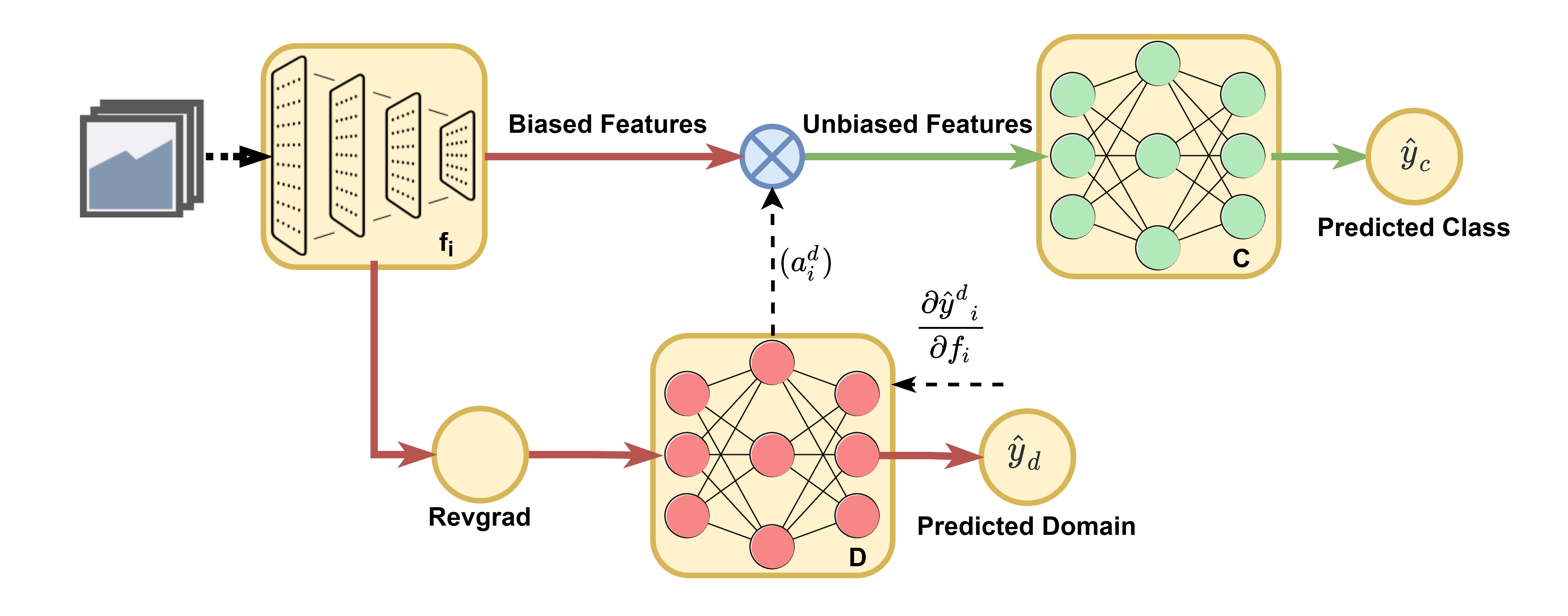}
       \caption{Illustration of training a adversarial model using proposed GBA (Gradient-based Activation) framework for debiasing. The features that are used by the classifier are filtered by discriminator's GBA, which uses discriminator's knowledge to inhibit domain discriminative features. Here the RevGrad is the gradient reversal layer \cite{ganin2015unsupervised}, dashed arrows represent the computation done during the backward propagation.}
      \label{fig:main}
 \end{figure*}

\noindent In Fig.~\ref{fig:disc_gba} we show the discriminator test accuracy plot of our approach, we see how using the proposed strategy provides the debias (near zero bias conflicting accuracy). On the bias aligned samples, the discriminator retains its accuracy. This is because, these are aligned with classification and the masking ensures that these features are not degraded during the adversarial training. This approach also ensures that we obtain high classification accuracy for both the bias aligned as well as bias conflicting samples as we show later. % without conflicting with the class correlations of the discriminator. It is worth noting how aligned accuracy is increasing steadily as compared to the other plots, this indicates learning from class correlation rather than bias.

\section{Proposed Approach}

\label{prop_app}

In this section we explain the proposed approach in detail, as illustrated in Fig.~\ref{fig:main}. 
\label{prop}

\subsection{Problem Formulation}

We formulate the problem of classification as a supervised learning problem, where the objective is to predict the class label $y^c$ for a given input $x$. The constraint is that the output variable must be unbiased with respect to some bias attribute variable $y^d$. Assume that there are total $N$ training samples of distribution $\mathcal{D}$, in form of tuples $\{x_i,y^c_i,y^d_i\}_{i=1}^N$ are available. Where $x_i$ is the input images, $y^c_i$ is the class label and $y^d_i$ refer to the bias attribute, independent of $y^c_i$. The predictor network $F(x)$ has access to the input variable $x_i$ and the bias attribute $y^d_i$. We follow the adversarial learning-based framework to debias towards the bias attributes while predicting the class labels. In adversarial learning,  a discriminator is trained to predict the bias attribute and a feature extractor is trained adversarially to it using a gradient reversal layer~\cite{ganin2015unsupervised}. This is clarified further in the next sub-section.

\subsection{Adversarial Learning for Bias Mitigation}

The adversarial learning framework consists of a feature extractor ($F$) and a  classifier or label predictor network ($C$). For incorporating adversarial learning, we use a discriminator network ($D$).  Feature extractor, classifier and discriminator are parameterised by parameters $\theta_f$, $\theta_c$ and $\theta_d$ respectively. 
The features $f_i$ of the input image ($x_i$) are encoded using the feature extractor, and these are classified by the classifier. The features are also provided to the discriminator to predict the bias attributes. The corresponding equations are given by:
\begin{equation}
 f_i =  F(x_i;\theta_f); \quad \hat{y}^c_i =  C(f_i,\theta_c); \quad \hat{y}^d_i =  D(f_i,\theta_d)
 \label{eq:1}
\end{equation}

% \begin{equation}
%  \hat{y}^c_i =  C(f_i,\theta_c)
%  \label{eq:2}
% \end{equation}

% \begin{equation}
%  \hat{y}^d_i =  D(f_i,\theta_d)
%  \label{eq:3}
% \end{equation}
 \vspace{-2em}
\begin{equation}
    \mathcal{L}_{c}=\frac{1}{N} \sum_{x_i \in \mathcal{D}} \mathcal{L}(\hat{y}^c_{i},y^c_i) \quad \mathcal{L}_{d}=\frac{1}{N} \sum_{x_i \in \mathcal{D}} \mathcal{L}(\hat{y}^d_{i},y^d_i)
\end{equation}
% \begin{equation}
%     \mathcal{L}_{d}=\frac{1}{N} \sum_{x_i \in \mathcal{D}} \mathcal{L}(\hat{y}^d_{i},y^d_i)
%     \label{eq:15}
% \end{equation}
where $N$ is the total number of images and $\mathcal{L}$ is the cross-entropy loss. In an adversarial learning setup, the total cost is obtained as:
\begin{equation}
    \mathcal{C}= \mathcal{L}_{c} -\lambda*\mathcal{L}_{d}
    \label{eq:16}
\end{equation}
\begin{equation*}
    (\hat{\theta}_f, \hat{\theta}_c) =\underset{\begin{subarray}{c}
  \theta_f, \theta_c \\
  \end{subarray}}{\text{arg min }} \mathcal{C}(\theta_f, \theta_c, \theta_{d})
\end{equation*}

\begin{equation*}
    (\hat{\theta}_d) =\underset{\begin{subarray}{c}
  \theta_d \\
  \end{subarray}}{\text{arg max }} \mathcal{C}(\theta_f, \theta_c, \theta_{d})
\end{equation*}

\noindent This is the standard setup for adversarial training. In the following subsections we consider our approach in more detail.

% \begin{equation}
%     \mathcal{L}_{c}=\frac{1}{N} \sum_{x_i \in \mathcal{D}} \mathcal{L}(\hat{y}_i,y_i)
% \end{equation}
% \begin{equation}
%     \mathcal{L}_{d}=\frac{1}{N} \sum_{x_i \in \mathcal{D}} \mathcal{L}(\hat{d}_i,d_i)
% \end{equation}

\subsection{Gradient Based Activations to Debias the Features}

Our approach towards bias mitigation is based on use of gradient based activations (GBA). The gradient of the output variable with respect to the input feature and its positive activation has been successfully applied to obtain  explainability about the prediction~\cite{selvaraju2017grad}. We follow a similar approach to identify the features used by discriminator in prediction. \\
The features obtained from Eq.\ref{eq:1} are entangled, i.e., contain both class and bias attribute information. As a result, the classifier minimizes the label prediction loss by considering both these features. For unbiased classification, the prediction must be independent of the bias attribute. In order to debias the classifier, we seek the discriminator's knowledge to identify the bias discriminative features and selectively mask the desired features to pass through the classifier. \\
We obtain the gradients of the prediction of the discriminator $\frac{\partial{\hat{y}}}{\partial{f_i}}$ w.r.t the features using the backward propagation to obtain features attended by discriminator. We mask the features with positive gradients when the discriminator is correct, and propagate the features with positive gradient when the discriminator is incorrect.\\
We define indicator variable $ind^d$  using the following condition in Eq \ref{eq:4}. $\hat{y}$ in the Eq \ref{eq:20} is the maximally activated logit, we obtain its gradients with respect to the features in Eq \ref{eq:5}. These gradients are then conditionally used to create the final mask vector $a^d_i$ in Eq \ref{eq:6}.

\begin{equation}
    ind^d = 
         \begin{cases}
           \text{1,} &\quad\text{if }\text{argmax(}\hat{y}^d_{i}\text{)}=y^d_{i} \\
           \text{-1} &\quad\text{otherwise} \\
         \end{cases}
         \label{eq:4}
\end{equation}

\begin{equation}
    \hat{y} = max({\hat{y}^d}_{i})   \label{eq:20}
\end{equation}

\begin{equation}
    g^d_i = \frac{\partial{\hat{y}}}{\partial{f_i}}
    \label{eq:5}
\end{equation}

\begin{equation}
 a^d_i = \begin{cases}
    0, & \text{if $(g^d_i.ind^d) $}>0 .\\
    1 & \text{if $ (g^d_i.ind^d)\le $} 0 
  \end{cases}
  \label{eq:6}
\end{equation}

The effective features for the classifier are obtained as follows:
\vspace{-1em}
\begin{equation}
    f_i^{cls}=f_i*a^d_i
    \label{eq:7}
    % \vspace{-1em}
\end{equation}
'$*$' represents the element-wise multiplications.

\noindent$f_i^{cls}$ is now used for the training in the adversarial manner as discussed in previous section.

\begin{table*}[]
 \centering
\scalebox{0.8}{
\begin{tabular}{|c|c|c|c|c|c |}
\hline
                                      &                                  & \multicolumn{3}{c|}{\textbf{Accuracy ($\uparrow$)}}           &                                                                               \\ \cline{3-5}
\multirow{-2}{*}{\textbf{Model Name}} & \multirow{-2}{*}{\textbf{Model}} & Aligned        & Conflicting    & Mean           & \multirow{-2}{*}\textbf{\begin{tabular}[c]{@{}l@{}}Bias\\ (GAP)($\downarrow$)\end{tabular}} \\ \hline
\textbf{Baseline}                     & N-way Softmax                    & {94.75 $\pm$ 0.25}         &{82.30 $\pm$ 0.31}         & 88.43 $\pm$ 0.20         & 12.45 $\pm$ 0.40                                                                                                  \\ 
\textbf{LfF\cite{nam2020learning}}                   & N-way Softmax                  & {90.33 $\pm$ 1.80}         & {68.64 $\pm$ 1.71}         & 79.49 $\pm$ 1.24         &  21.69 $\pm$ 2.48                                                                                                   \\ 
\textbf{Domain Independent\cite{wang2020towards}}           & N-way Classifier per Domain      &{92.38 $\pm$ 0.20}          & {91.86 $\pm$ 0.21}         & \textbf{{92.12} $\pm$ 0.15}          &  0.52 $\pm$ 0.29                                                                                                 \\ \hline
\textbf{Adversarial}                  & Gradient Reversal                & {86.98 $\pm$ 0.70}         & {86.61 $\pm$ 0.37}          & 86.80 $\pm$ 0.40         & 0.37 $\pm$ 0.80                                                                                                  \\ 

\textbf{Adversarial with GBA}         & Proposed                         & {91.95 $\pm$ 0.22} & \textbf{{92.05 $\pm$ 0.23}} & {92.00 $\pm$ 0.15} &
\textbf{0.10 $\pm$ 0.31}                                                          \\ \hline
\end{tabular}
}
\vspace{-0.5em}
\caption{Performance comparison of different algorithms on CIFAR-10S, Here we show test accuracy on the bias aligned and bias conflicting samples as a measure of biasness, It can be seen that GBA is the best in terms of debiasing}
   \label{tab:cifar-s}
\end{table*}
% \vspace{-2em}
\begin{table*}[h!]
 \centering
 \scalebox{0.8}{
\begin{tabular}{|c|c|c|c|c|c |}
\hline
                                      &                                  & \multicolumn{3}{c|}{\textbf{Accuracy ($\uparrow$)}}           &                                                                                \\ \cline{3-5}
\multirow{-2}{*}{\textbf{Model Name}} & \multirow{-2}{*}{\textbf{Model}} & Aligned        & Conflicting    & Mean           & \multirow{-2}{*} \textbf{\begin{tabular}[c]{@{}l@{}}Bias\\ (GAP)($\downarrow$)\end{tabular}} \\ \hline
\textbf{Baseline}                     & N-way Softmax                    & {87.94 $\pm$ 0.36}          & {69.39 $\pm$ 0.42}         & 78.67 $\pm$ 0.27          & 18.55 $\pm$ 0.55     \\                                   \textbf{LfF\cite{nam2020learning}}               & N-way Softmax                    & {87.01 $\pm$ 0.63}         & {56.87 $\pm$ 0.72}         & 71.93 $\pm$ 0.47          & 30.14 $\pm$ 0.95        \\                                                           \textbf{Domain Independent\cite{wang2020towards}}        & N-way Classifier per Domain      & {88.39 $\pm$ 0.15}         & {78.14 $\pm$ 0.13}        & 83.26 $\pm$ 0.01          & 10.25 $\pm$ 0.20                                                                                                \\ \hline

\textbf{Adversarial}                  & Gradient Reversal                & {85.52 $\pm$ 0.65 }       & {75.74 $\pm$ 0.29}         & 80.63 $\pm$ 0.35         & 9.78 $\pm$ 0.71                                                                                               \\ 

\textbf{Adversarial with GBA}         & Proposed                         & {88.81 $\pm$ 0.19} &\textbf{{79.46 $\pm$ 0.22}} & \textbf{{84.41} $\pm$ 0.14} & \textbf{9.35 $\pm$ 0.29}                                                                   \\ \hline
\end{tabular}
}
\vspace{-0.5em}
\caption{Performance comparison on a non-linear transformation, CIFAR-I setting, here also our algorithm outperforms existing algorithms in both bias and accuracy}
    \label{tab:cifar-i}
    \vspace{-2em}
\end{table*}

\section{Experiments and Results}
\label{results}

% \subsection{Bias Metric} 
%  We follow similar approach as discuss in [Wang \textit{et al., 2020}] and Zhao\textit{ et al} to define the bias metric.
% \begin{equation}
%     \frac{1}{|C|} \sum_{c \in C} \frac{\max \left(\mathrm{Gr}_{c}, \mathrm{Col}_{c}\right)}{\mathrm{Gr}_{c}+\mathrm{Col}_{c}}-0.5
% \end{equation}
% $\mathrm{Gr}_{c}$ and $\mathrm{Col}_{c}$ are the predicted samples for a class $c$ for both the greyscale and color domains respectively.From the given equation we can observe that, for the low bias case, the accuracy for individual domains should be highly correlated, thus only this metric is not sufficient measure to analyse the bias. Further, one should note that the synthetic datasets (variations of CIFAR-10 and CMNIST) are constructed in such a way, so that there is no trade-off between accuracy and fairness as the train set is highly skewed by domain, while the test set is balanced with respect to the same. So, accuracy is a good metric for such tasks[Wang \textit{et al., 2020}]. However, the bias metric defined above is not the complete indication of a model's fairness, so according to us, mean accuracy is the primary metric, and the bias metric was provided to remain consistent with the literature.
\begin{table}[]
	\scalebox{0.85}{
\centering
\begin{tabular}{|c|c|c|}
\hline
\multirow{2}{*}{\textbf{Model}} & \multicolumn{2}{c}{\textbf{Accuracy}}                                               \\ \cline{2-3} 
                                & \textbf{95 \% Skew}                       & \textbf{98 \% Skew}                      \\ \hline
\textbf{Vanilla}                & 77.63 $\pm$ 0.44                          & 62.29$\pm$ 1.47    \\                      
\textbf{Domain Ind\cite{wang2020towards} }     & 65.82$\pm$0.81                            & 45.39$\pm$1.20    \\  
\textbf{Filter-Drop\cite{nagpal2020attribute}}                    & 78.44$\pm$0.58                            & {62.31$\pm$1.72} \\
\textbf{Group-DRO\cite{Sagawa2020Distributionally}}                    & 84.50 $\pm$0.46                           & 76.30$\pm$1.53 \\
\textbf{REPAIR\cite{li2019repair}}                    & {82.51$\pm$0.59}                           & {72.86$\pm$1.47} \\
\textbf{LfF\cite{nam2020learning}}                    & 85.39$\pm$0.94                            & \textbf{80.48$\pm$0.45} \\ \hline 
\textbf{Adversarial}            & 80.35$\pm$0.52                            & 64.83 $\pm$ 0.34               \\
\textbf{Adversarial with GBA}   & \textbf{87.92 $\pm$ 0.6} & 79.11 $\pm$ 1.6  
\\  \hline 
\end{tabular}
}
\vspace{-0.5em}
\caption{ Performance comparisons in term of classification accuracy on Colored MNIST dataset.}
	\label{tab:colormn}
	\vspace{-0.5em}
\end{table}

\subsection{Datasets}
We evaluate the proposed model on the following standard datasets : 
\textbf{CIFAR-10S~\cite{wang2020towards}:} It is a skewed  version of CIFAR-10~\cite{darlow2018cinic}, presented by Wang~\textit{et al}~\cite{wang2020towards}. This data contains tranformational bias. It consists of 50,000 images of size 32×32 of 10 object classes. Each class has a total of 5000 images. CIFAR-10S is divided into two domains color and greyscale domains. In this datasets per class, the 50,000 training images are split 95\% to 5\% between the two domains; five classes are 95\% color, and five classes are 95\% greyscale. For testing we evaluate each class on bias aligned and conflicting samples. \\
\textbf{CIFAR-I:} It is an extension of CIFAR-10S~\cite{wang2020towards}, where the images of the skewed domain are taken from similar classes of ImageNet~\cite{imagenet_cvpr09}. So in this dataset, there are 10 classes and two domains (bias attributes).\\
\textbf{ColoredMNIST:} The ColoredMNIST dataset containing colour bias is taken from~\cite{nam2020learning}. In this dataset, images of greyscale MNIST dataset are injected  10 colors with random perturbation in each class, resulting in a dataset with 10 classes of digits and 10 domains of colors. \\
\textbf{CelebA:} It is a real world multi-attribute dataset consisting of 40 attributes for each image. Here we have an example of real world bias in form of Gender. We find that Hair Color attribute and Heavy Makeup are the most correlated to the bias attribute (Gender) as done by \cite{nam2020learning}, so we perform experiment on two setups, Hair Color as target attribute and Gender as bias attribute and Heavy Makeup as target attribute and Gender as bias attribute.

\subsection{Training Setup}
 For the CIFAR-10S, CIFAR-I and CelebA datasets, we use the Resnet-18~\cite{he2016deep} model, where the last fully connected layer is replaced with two consecutive fully connected layers. In the Colored MNIST dataset, we use the multi-layered perceptron consisting of three hidden layers as the feature extractor.
 %commented
%  The code and other training details are provided in the supplementary material.
 
 %We use the SGD optimizer with batch size 128, classifier learning rate of 0.1 with decay, and a constant discriminator learning rate of 1e-4 for CIFAR-10S and CIFAR-I datasets. In the ColorMNIST dataset, we use Adam optimizer with batch size 256, classifier learning rate 1e-3, discriminator learning rate of 1e-5 and $\lambda$ 0.1 for 95$\%$ skew and 0.01 for 98$\%$. The code and other details are provided in the supplementary material.
\subsection{Results and Discussion}
\label{results}

\subsubsection{CIFAR-10S}

We use a ResNet-18~\cite{he2016deep} model trained on CIFAR-10S as the vanilla baseline. The adversarial model is using the gradient reversal layer as in~\cite{ganin2015unsupervised}. We evaluate the methods on bias aligned and bias conflicting accuracies along with their mean.
For the unbiased model both these accuracies must be close. We measure this using \textit{bias gap} which is the difference between the aligned and conflicting  accuracies. In Fig \ref{fig:clf_all} we show the bias aligned and bias conflicting accuracy plots on different methods, bias gap of baseline vanilla model in Fig \ref{fig:graph_base} highlights the problem of bias in standard deep learning models. Fig \ref{fig:graph_adv} show how adversarial methods can reduce the bias gap as compared to baseline but degrades the overall class accuracy. Fig \ref{fig:graph_dnew} show the unbiased learning of the proposed approach, note that it has solved the problem of bias-accuracy trade-off which was there in adversarial methods.

Further we show performance comparison with baselines and recent techniques in Table \ref{tab:cifar-s}. We observe a increase of \textbf{3.5\%} and \textbf{5.5\%} in accuracy as compared to the vanilla baseline and adversarial approach respectively along with greater reduction in bias. Moreover we observe that the proposed approach also outperforms other recent works in the task of bias mitigation.  

\begin{table}[!h]
		\centering
\begin{tabular}{|c|c|c|c|c|}
\hline
\multirow{2}{*}{\textbf{Model}} & \multicolumn{3}{c|}{\textbf{Accuracy}}           & \multirow{2}{*}{\begin{tabular}[c]{@{}l@{}}Bias \\ (GAP)\end{tabular}} \\ \cline{2-4}
                                & Aligned        & Conflicting    & Mean           &                                                                        \\ \hline
GBA-n                           & 92.05          & 90.60          & 91.32          & 1.45                                                                   \\ \hline
GBA-c                           & 90.69          & 90.06          & 90.37          & 0.63                                                                   \\ \hline
GBA                             & 92.05 & \textbf{91.95} & \textbf{92.00} & \textbf{0.10}                                                          \\ \hline
\end{tabular}
\caption{Results on the ablation of various activations on CIFAR-10S datasets.}
    \label{tab:GBA_abl}
    \vspace{-1em}
\end{table}

\begin{table*}[]
\scalebox{0.8}{
\begin{tabular}{|c|c|c|c|c|c|c|c|c|}
\hline
\multirow{2}{*}{\textbf{Model}} & \multicolumn{4}{c|}{\textbf{Heavy Makeup}}                                  & \multicolumn{4}{c|}{\textbf{Hair Color}}                                    \\ \cline{2-9} 
                                & \textbf{Aligned} & \textbf{Conflicting} & \textbf{Mean} & \textbf{Bias Gap} & \textbf{Aligned} & \textbf{Conflicting} & \textbf{Mean} & \textbf{Bias Gap} \\ \hline
Vanilla                        & 92.44 ± 0.74     & 31.46 ± 2.45         & 61.95 $\pm$ 1.28         & 60.98 $\pm$ 2.56            & 90.58 ± 0.34     & 57.35 ± 0.21         & 73.97 $\pm$ 0.20        & 33.23 $\pm$ 0.40            \\ 
LfF~\cite{nam2020learning}                             & 83.85 ± 1.68     & 45.54 ± 4.28         & 64.69 $\pm$ 2.29       & 38.31 $\pm$ 4.60            & 88.85 ± 1.27     & 80.24 ± 2.16         & 84.55 $\pm$ 1.25        & 8.61 $\pm$ 2.5             \\

Domain Independent~\cite{wang2020towards}                             & 79.88$\pm$1.71     & 43.24$\pm$4.33         & 61.56$\pm$2.31       & 36.64 $\pm$ 5.64            & 90.97$\pm$3.71     & 79.25$\pm$3.33         & 85.11$\pm$2.67        & 7.44 $\pm$ 3.21             \\

Group DRO~\cite{Sagawa2020Distributionally}                             & 79.28$\pm$1.20     & 46.24$\pm$3.61         & 62.76$\pm$2.22        & 33.04 $\pm$ 3.22            & 89.68$\pm$0.65     & 81.41$\pm$1.47         & 85.55$\pm$0.88        & 8.27 $\pm$ 2.01             \\

\hline

Adversarial            & 92.07 ± 2.88     & 33.79 ± 3.81         & 62.93 $\pm$ 2.38        & 58.28 $\pm$ 4.77             & 93.4 ± 0.91      & 62.75 ± 3.47         & 78.08 $\pm$ 1.79        & 30.65 $\pm$ 5.59            \\ 
Adversarial with GBA                             & 81.49 ± 1.91     & \textbf{49.79± 3.15}          & \textbf{65.64 $\pm$ 1.55}         & \textbf{31.70 $\pm$ 3.10}              & 90.67 ± 1.01     & \textbf{83.28 ± 1.83}         & \textbf{86.98 $\pm$ 1.04}         & \textbf{7.39 $\pm$ 2.09}              \\ \hline
\end{tabular}
}
    \caption{Results on the Heavy Makeup and Hair Color Attributes of the CelebA Dataset, with the bias attribute being the gender, here we show that the Simple Adversarial method was unable to debias the model, the representation. Using GBA, we reduce the bias gap greatly when compared to the Adversarial method, and while maintaining state of the art accuracy.}
 \label{tab:celeba}
\end{table*}

\subsubsection{CIFAR-I}

The color-greyscale transformation of CIFAR-10S is one difference in terms of data distribution. Another case of bias could be in terms of distribution of data samples. We evaluate our algorithm on such a transformation to simulate real-world data using CIFAR-I with samples from another datasets. The other dataset we use is that of similar classes from ImageNet dataset. This distribution change exhibits a different  bias as compared to the color-greyscale transformation. In Table~\ref{tab:cifar-i} we report and compare the performance of the proposed method with baselines and recent techniques. We can observe that the proposed method achieves a boost of approximately\textbf{ 6\%}  and \textbf{4\%} in mean accuracy over the vanilla baseline and adversarial methods respectively and outperforms the state-of-the-art domain independent approach in both mean accuracy and bias removal.

\subsubsection{Colored MNIST}
Another case of biased learning is in Colored MNIST dataset~\cite{nam2020learning}, where MNIST dataset is injected with colors for each class respectively. In this case, the neural network generally learns to classify them on the basis of color rather than learning about digits. The previous two datasets had only two bias attributes to discriminate; in this dataset we have ten bias attributes, using this dataset we test the scalability of our model to multiple number of attributes. The performance on this dataset has been reported in Table~\ref{tab:colormn} for different level of skews. Here again we see along with outperforming the recent methods the proposed model improves on the adversarial method by a large number. We note the recent domain independent~\cite{wang2020towards} network performs poorly in this multiple domains setting.

% \begin{wrapfigure}{r}{0.5\linewidth}
% \centering
% \includegraphics[height=6cm,width=8cm]{Journal-JSTSP/fig/skew.png}
%       \caption{Comparison of average accuracy between the algorithms at different levels of skew in CIFAR-10S dataset}
% \label{fig:skew}
% \end{wrapfigure}

% \begin{figure}[h!]
%  \centering
%     \includegraphics[height=6cm,width=8cm]{Journal-JSTSP/fig/skew.png}
%       \caption{Comparison of average accuracy between the algorithms at different levels of skew in CIFAR-10S dataset}
%       \label{fig:skew}
%  \end{figure}

% We tested our approach for varying level of training skews on CIFAR-10S dataset and compared it with other algorithms.In Fig \ref{fig:skew} we see sharp decline in accuracy of baseline with increase in skew. ADV . GBA greatly improves the simple adversarial method and is competitive with the state-of-the-art Domain Independent \cite{wang2020towards} approach even on higher skews. 

\begin{figure*}[h!]
\centering
\begin{subfigure}{0.33\textwidth}
  \centering
  \includegraphics[width=1.0\linewidth]{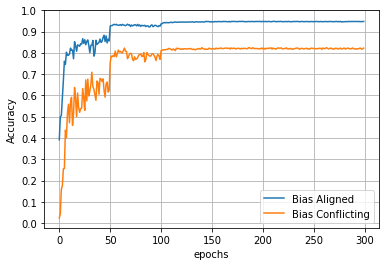}
  \vspace{-1.5em}
  \caption{Baseline}
  \label{fig:graph_base}
\end{subfigure}
\begin{subfigure}{0.33\textwidth}
  \centering
  \includegraphics[width=1.0\linewidth]{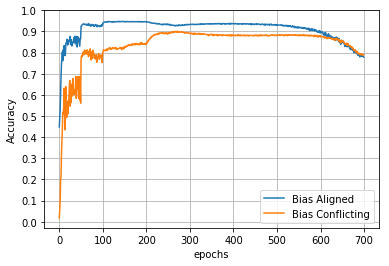}
  \vspace{-1.5em}
  \caption{Adversarial}
  \label{fig:graph_adv}
\end{subfigure}
\begin{subfigure}{0.33\textwidth}
  \centering
  \includegraphics[width=1.0\linewidth]{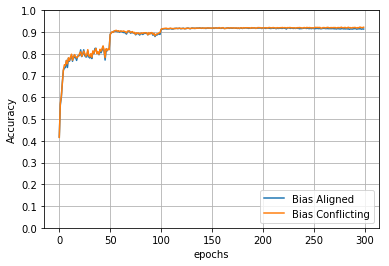}
  \vspace{-1.5em}
  \caption{GBA }
  \label{fig:graph_dnew}
\end{subfigure}
\vspace{-1em}
\caption{Figures above show the validation accuracy of bias aligned and bias conflicting samples over the course of training on CIFAR-10S dataset. We observe that the baseline model has poor performance on the bias conflicting samples compared to the bias aligned samples. The adversarial model improves upon the baseline, but the trade-off is evident as to completely debias the model, the class features are harmed. GBA with the adversarial framework makes it completely fair in terms of the bias, there is almost no discrepancy in the aligned and conflicting accuracy, and the average accuracy also improves significantly.}
 \label{fig:clf_all}
 \vspace{-0.5em}
\end{figure*}

\begin{figure*}[h!]
\centering
\begin{subfigure}{0.45\textwidth}
  \centering
  \includegraphics[width=0.8\linewidth]{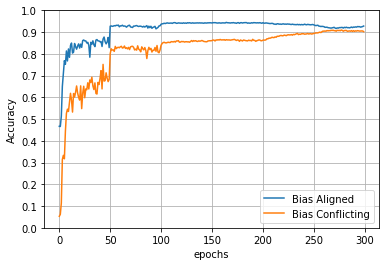}
  \vspace{-0.5em}
  \caption{GBA-n, Naive masking using GBA}
  \label{fig:GBA-n}
\end{subfigure}
\begin{subfigure}{0.45\textwidth}
  \centering
  \includegraphics[width=0.8\linewidth]{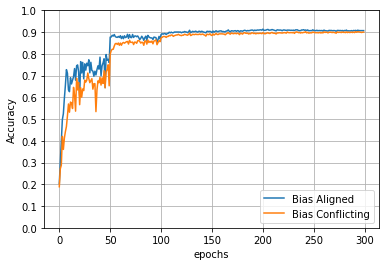}
  \vspace{-0.5em}
  \caption{GBA-c, Only correct masking using GBA}
  \label{fig:GBA-c}
\end{subfigure}
\vspace{-1em}
\caption{Ablation study of masking rule by GBA on CIFAR-10S dataset.}
 \label{fig:GBA_abl}
  \vspace{-0.5em}
\end{figure*}

\subsubsection{CelebA}
This dataset contains gender bias with respect to the heavy makeup and hair color attribute. To evaluate different algoritms on these attribute learning tasks we report the accuracy on bias aligned and bias conflicting samples, along with mean accuracy and bias gap for particular target attribute on the unbiased test set in Table \ref{tab:celeba}. We observe adversarial method improving by \textbf{3\%} and \textbf{9\%} in average accuracy and \textbf{7\%} and \textbf{1.2\%} in bias gap on heavy makeup and hair color attributes respectively. We also observe performance of the proposed method outperforms the recent state-of-the-art methods like LfF~\cite{nam2020learning}.
\\
In this section we discussed the performance of different methods on various datasets and metrics. We observe GBA greatly improves the accuracy of adversarial method, moreover the proposed approach is the best performing method averaged across all the datasets. In the results discussed above adversarial with GBA is on average \textbf{5\%} better than the second best method-LfF in terms of accuracy. The other details and experiments are provided in the project page~\footnote{https://vinodkkurmi.github.io/GBA/}.\\

\vspace{-2em}
\section{Analysis}
\label{analysis}
\begin{figure}[h!]
\vspace{-1em}
\centering
\begin{subfigure}{.24\textwidth}
  \centering
  \includegraphics[height=2cm,width=4cm]{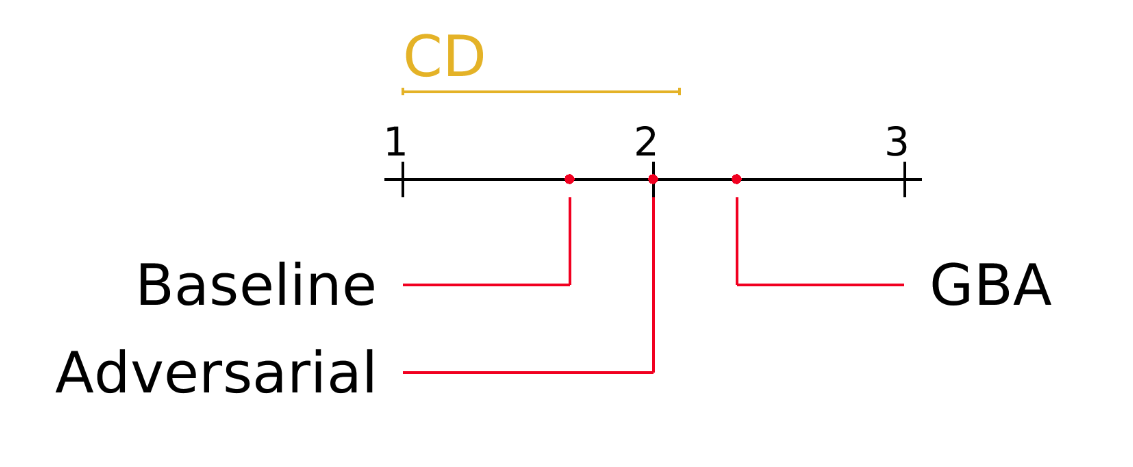}
   \vspace{-1em}
  \caption{}
    \vspace{-1em}
  \label{fig:ssa_cifar}
\end{subfigure}%
\begin{subfigure}{.24\textwidth}
%   \centering
  \includegraphics[height=2cm,width=4cm]{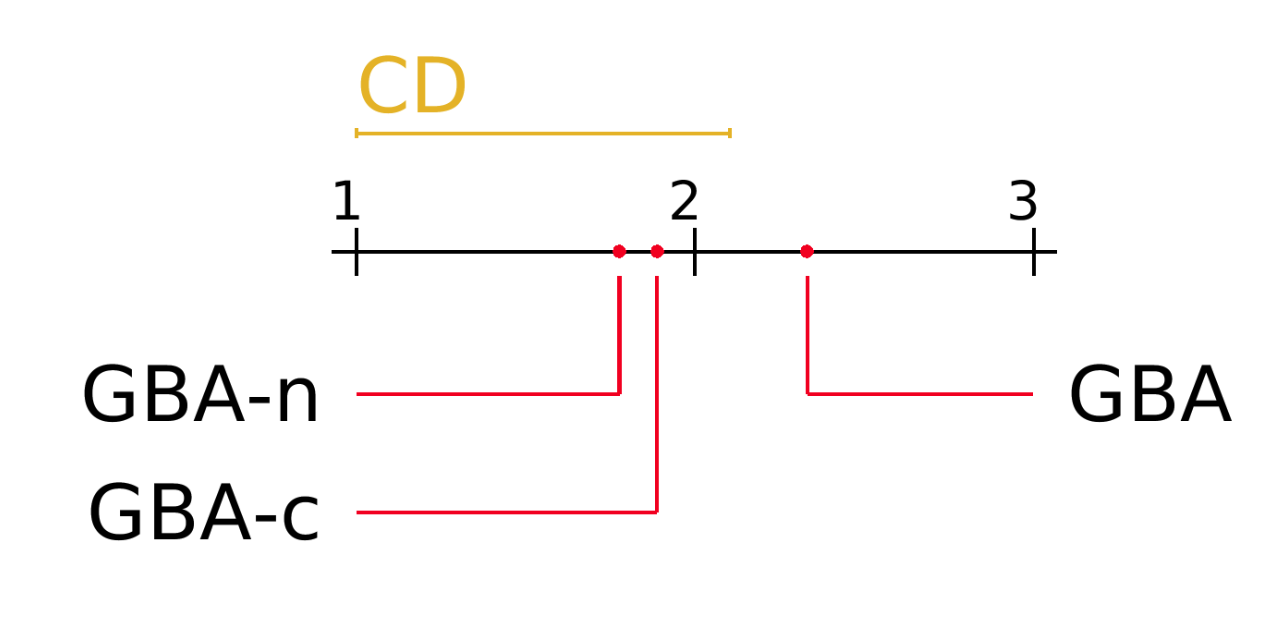}
 \vspace{-1em}
  \caption{}
%  \vspace{-1em}
  \label{fig:ssa_abl}
\end{subfigure}
\caption{Statistical significant test for CIFAR-10S datasets on (a)  baseline, adversarial, and  GBA method  (b) different variations of proposed model GBA-c, GBA-n and  GBA. }
 \vspace{-1em}
\end{figure}

\subsection{Ablation Study of Gradient Based Activations}
\label{abl:gba}

In the proposed approach of gradient based activations (GBA), masking rule for the features is conditioned on discriminator's prediction, in this section we analyse how this conditioning is crucial for the performance of GBA. Fig \ref{fig:GBA-n} represents bias aligned and conflicting accuracy if we naively mask features attended by discriminator (GBA-n) i.e. drop the features that our discriminator is using for prediction without considering whether the discriminator is correct or incorrect. The plot when compared to Fig~\ref{fig:graph_dnew} shows highly biased trend highlighting the importance of right conditioning.\\
To improve upon GBA-n, in Fig~\ref{fig:GBA-c} we see the variation GBA-c where we only attend to the classifier when the discriminator correctly classifies the input and pass the raw un-attended features when the discriminator is incorrect. Here we can see better bias removal than GBA-n version which support our hypothesis that discriminator correlates with the class features to predict the incorrect domain. Hence, enhancing these features while training promotes unbiased learning and improves class prediction ability as seen in Fig \ref{fig:graph_dnew}. In Table \ref{tab:GBA_abl}, we report the performance of different ablations where we see a systematic improvement in bias and accuracy as we apply different components of GBA.

\subsection{Statistical Significance Analysis}
We analyze the statistical significance~\cite{demvsar_JMLR2006} for the proposed method in bias mitigation for CIFAR-10S dataset. The Critical Difference (CD) is related to the confidence level (it is 0.05 in our case) for the number of tested datasets and average ranks. If the methods' rank difference is outside the CD (it is 1.048 for our case), it implies that these two methods are significantly different. 
In Fig.~\ref{fig:ssa_cifar} and Fig~\ref{fig:ssa_abl}, We  provide the statistical test for baselines, adversarial with the proposed method and different variations of the proposed method defined in the previous section. It visualizes the post hoc analysis using the CD diagram for CIFAR-10S dataset. From the figures, it is clear that the proposed method is significantly different from the baseline model and adversarial method.

\vspace{-1em}
\section{Conclusion}

Through this work, we provide a method to address the crucial problem in the adversarial learning framework to obtain unbiased classification. Through extensive empirical analysis on multiple standard datasets we show that the proposed approach works well. Specifically, we showed that gradient based activation uses a biased discriminator's gradients in order to debias the classifier. Our ablation analysis also justifies the use of the proposed method. Through our work, we also obtain a better understanding of debiasing a classifier, particularly in an adversarial setting.

{\small
\bibliography{aaai22}
}
\end{document}